# Quantifying the pathways to life using assembly spaces

Stuart M. Marshall,[1] Douglas G. Moore,[2] Alastair R. G. Murray,[1] Sara I. Walker,[2,3*] and Leroy Cronin[1*]

[1] School of Chemistry, University of Glasgow, Glasgow, G12 8QQ, UK.

[2] BEYOND Center for Fundamental Concepts in Science, Arizona State University, Tempe, AZ, USA

[3] School of Earth and Space Exploration, Arizona State University, Tempe, AZ, USA

*Corresponding author email: Lee.Cronin@glasgow.ac.uk, sara.i.walker@asu.edu

**Abstract**

We have developed the theory of pathway assembly to explore the extrinsic information required to distinguish a given object from a random ensemble. To quantify the assembly in an agnostic way, we determine the pathway assembly information contained within such an object by deconstructing the object into its irreducible parts, and then evaluating the minimum number of steps to reconstruct the object. The formalisation of this approach uses an assembly space. By finding the minimal number of steps contained in the route by which the objects can be assembled within that space, we can compare how much information ($I$) is gained from knowing this pathway assembly index (PA) according to $I_{PA} = log \frac{|N|}{|N_{PA}|}$ where, for an end product with $PA = x$, N is the set of objects possible that can be created from the same irreducible parts within $x$ steps regardless of PA, and $N_{PA}$ is the subset of those objects with the precise pathway assembly index $PA = x$. Applying this theory to objects formed in 1D, 2D and 3D leads to the identification of objects in the world or wider Universe that have high assembly numbers. We propose that objects with high PA will be uniquely identifiable as those that must have been produced by biological or technological processes, rather than the assembly occurring via unbiased random processes alone, thereby defining a new scale of aliveness. We think this approach is needed to help identify the new physical and chemical laws needed to understand what life is, by quantifying what life does.

## Introduction

In the thought experiment known as the "infinite monkey theorem", an infinite number of monkeys, each having a typewriter, produce strings of text by hitting keys at random (1). Given infinite resources, it can be deduced that the monkeys will produce all possible strings, including the complete works of Shakespeare. However, when constrained to the bounds of the physical universe, the likelihood that any particular text is produced by a finite number of monkeys drops rapidly with the length of the text (2). This can also be extended to physical objects like cars, aeroplanes, and computers, which must be constructed from a finite set of objects - just as meaningful text is constructed from a finite set of letters. Even if we were to convert nearly all matter in the universe to object constructing monkeys, and give them the age of the universe in which to work, the probability that any monkey would construct any sufficiently complex physical object is negligible (3). This is an entropic argument – the number of possible arrangements of the objects of a given composition increases exponentially with the object size. For example, if the number of possible play-sized strings is sufficiently large, it would be practically impossible to produce a predetermined Shakespearean string without the author. This argument implies information external to the object itself is necessary to construct an object if it is of sufficiently high complexity (4,5): in biology the requisite information partly comes from DNA, the sequence of which has been acquired through progressive rounds of evolution. Although Shakespeare's works are – in the absence of an appropriate constructor (6) (an author) - as likely to be produced as any other string of the same length, our knowledge of English, and Shakespeare in particular, allows us to partition the set of possible strings to generate information about those strings containing meaning, and to construct them.

Biological systems have access to a lot of information - genetically, epigenetically, morphologically, and metabolically - and the acquisition of that information occurs via evolutionary selection over successive cycles of replication and propagation (7). One way to look at such systems is by comparing the self-dissimilarity between different classes of complex system, allowing a model free comparison (8). However, it has also been suggested that much of this information is effectively encrypted, with the heritable information being encoded with random keys from the environment (9). As such, these random keys are recorded as frozen accidents and increase the operative information content, as well as help direct the system during the process of evolution, producing objects that can construct other objects (10). This is significant since one important characteristic of objects produced autonomously by machinery (such as life), which itself is instructed in some way, is their relative complexity as compared to objects that require no information for their assembly beyond what chemistry and physics alone can provide. This means that for complex objects there is 'object-assembly' information that is generated by an evolutionary system, and is not just the product of laws of physics and chemistry alone. Biological systems are the only known source of agency in the universe (11), and it has been suggested that new physical laws are needed to understand the phenomenon of life (12). The challenge is how to explore the complexity of objects generated by evolutionary systems without *a priori* having a model of the system.

Herein, we present the foundations of a new theoretical approach to agnostically quantify the amount of potential pathway assembly information contained within an object. This is achieved by considering how the object can be deconstructed into its irreducible parts, and then evaluating the minimum number of steps necessary to reconstruct the object along any pathway. The analysis of pathway assembly is done by the recursive deconstruction of a given object using shortest paths, and this can be used to evaluate the effective pathway assembly index for that object (13). In developing pathway assembly, we have been motivated to create an *intrinsic* measure of an object forming through random processes, where the only knowledge required of the system is the basic building blocks and the permitted ways of joining structures together. This allows determining when an *extrinsic* agent or evolutionary system is necessary to construct the object, permitting the search for complexity in the abstract, without any specific notions of what we are looking for, thus removing the requirement for an external imposition of meaning, see Figure 1.

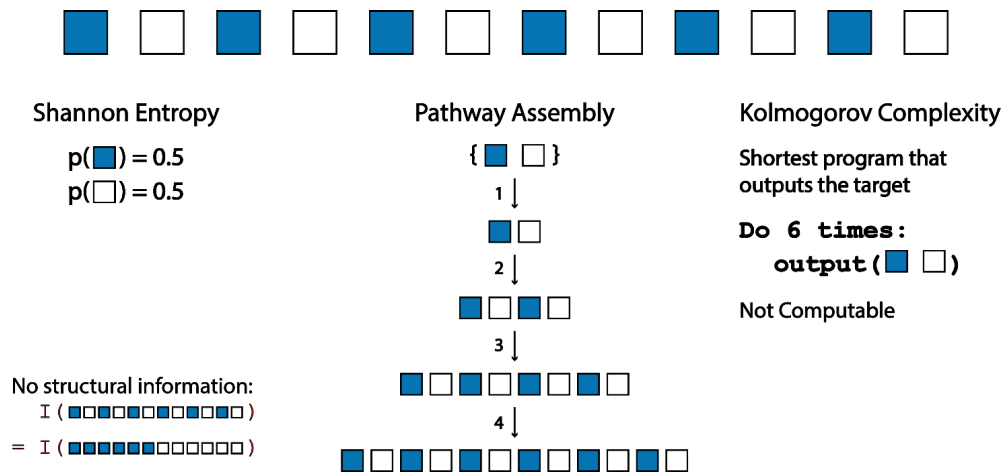

**Figure 1:** The Pathway Assembly process (centre) (13) is compared to implementations of Shannon Entropy (14) (left) and Kolmogorov Complexity (15) (right) for blue and white blocks. The Pathway Assembly process leads to a measure of structural complexity that accounts for the structure of the object and how it could have been constructed, which is in all cases computable and unambiguous.

The development of the Pathway Assembly (13) index (PA) was motivated by the desire to define a biological threshold, such that any object found in abundance with PA above the threshold would have required the intervention of one or more biological processes to form (16). The Pathway Assembly index (PA) of an object is the length of the shortest pathway to construct the object starting from its basic building blocks. It should be noted that this approach is entirely classical (17), allowing quantifying pathways through assembly space probabilistically as a way to understand what life does. We construct the object using a sequence of joining operations, where at each step any structures already created are available for use in subsequent steps, see Figure 2. The shortest pathway approach is in some ways analogous to Kolmogorov complexity (15), which in the case of strings is the shortest computer program that can output a given string. However, Pathway Assembly differs in that we only allow joining operations as defined in our model. This restriction is intended to allow the Pathway Assembly process to mimic the natural construction of objects through random processes, and it also importantly allows the PA of an object to be computable for all finite objects (see Theorem 4 in the SI).

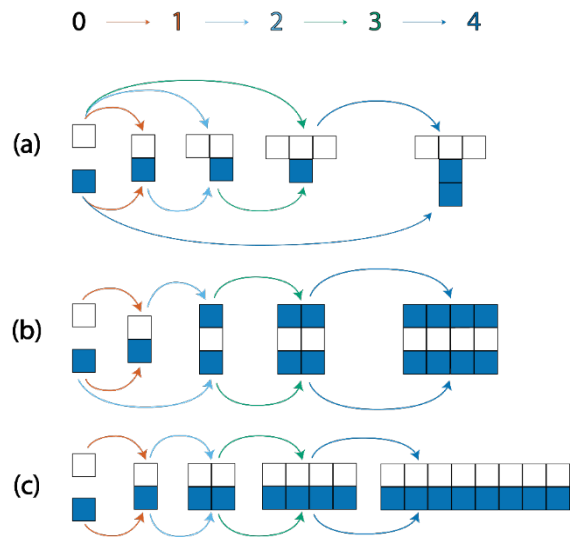

**Figure 2:** The basic concept of pathway assembly is shown here. Each of the final structures can be created from white and blue basic objects in four joining operations, giving a Pathway Assembly Index of 4. Pathway (a) shows the creation of a structure that can only be formed in four steps by adding one basic object at a time, while pathway (c) represents the maximum increase in size per step, by combining the largest object in the pathway with itself at each stage. Pathway (b) is an intermediate case.

Given a system where objects interact randomly and with equal probability, it is intuitively clear that the likelihood of an object being formed in $n$ steps decreases rapidly with $n$. However, it is also true that a highly contrived set of biases could guarantee the formation of any object. For example, this could occur if we were to model the system such that any interactions contributing to the formation of the object were certain to be successful, while other interactions were prohibited. For complex objects, such a serendipitous set of biases would seem unlikely in the absence of external information about the end products, but physical systems generally do have biases in their interactions, and we can explore how these affect the likelihood of formation of objects. However, we expect for any perceived "construction processes" that requires a large enough set of highly contrived biases, we can deduce that external information is required in the form of a "machine" that is doing the constructing.

Technological processes are bootstrapped to biological ones, and hence, by extension, production of technosignatures involves processes that necessarily have a biological origin. Examples of biosignatures and technosignatures include chemical products produced by the action of complex molecular systems such as networks of enzymes (18), and also objects whose creation involved any biological organisms such as technological artefacts (19), complex chemicals made in the laboratory (20), and the complete works of Shakespeare. Finding the object in some abundance, or a single object with a large number of complex, but precisely repeating features, is required in order to distinguish single random occurrences from deliberately generated objects. For example, a system which produces long random strings will generate many that have high PA, but not in abundance. Finding the same long string more than once will tell us that there is a bias in the system towards creating that string, thus searching for signatures of life should involve looking for objects with high PA found in relatively high abundance.

**Formalism**

In this manuscript, we explore the foundations of Pathway Assembly, as well as some of its properties and variants. We also describe how Pathway Assembly can be incorporated into a new information measure, Pathway Information, and how this can help identify objects, above a threshold, that must have been produced by living systems. Finally, we offer some examples of the use of pathway assembly in systems of varying dimensionality, and describe some potential real-world applications of this approach. The Pathway Assembly process is formally defined in the context of an Assembly Space, which comprises an acyclic quiver $\Gamma$ (a quiver being a directed graph that allows multiple edges between pairs of nodes and has no directed cycles), where the vertices in the quiver are objects in the space, along with an edge labelling map $\phi$ which associates each edge with a vertex in the quiver (see Definition 11 in the SI). The quiver is associated with a reachability relationship $\leq$ where for vertices $a, b$ in $\Gamma$, $a \leq b$ if there is a path from $a$ to $b$, in other words it is possible reach $b$ starting at $a$ by following a sequence of edges along their respective directions. If for an edge $e$ from object $x$ to object $z$, $\phi(e) = y$, then this can be thought of as $z$ being constructed through the combination of $x$ and $y$. We also require that the symmetric operation exists within the space, i.e. there is an edge $g$ from $y$ to $z$ such that $\phi(g) = x$.

We define an assembly subspace $\Delta$ on an assembly space $\Gamma$ to be an assembly space that contains a subset of the objects in $\Gamma$, maintaining all the relationships between them (see Definition 15 in the SI). An assembly subspace is said to be rooted if it contains a nonempty subset of the basic objects. This is an important distinction in the definition of the Assembly Index below, as it allows us to define the shortest construction pathway for objects using a consistent set of basic objects. We define the basis of an Assembly Space $\Gamma$ as the set of minimal vertices in the space with regard to $\leq$, and refer to those vertices as basic objects, basic vertices, or basic elements (see Definition 12 in the SI). We define an *assembly map* (see Definition 17 in the SI) as a map from one assembly space $\Gamma$ to another $\Delta$ that maintains the relationship between objects, but may map multiple objects in $\Gamma$ to the same object in $\Delta$. One such map that is generally applicable is the mapping of each object to its size, see Figure 3. Assembly maps can be useful for finding a lower bound to the assembly index (described below, and Definition 19 in the SI), by mapping to a system that may be more computationally tractable to work in than the original system of interest (see Theorem 3 in the SI).

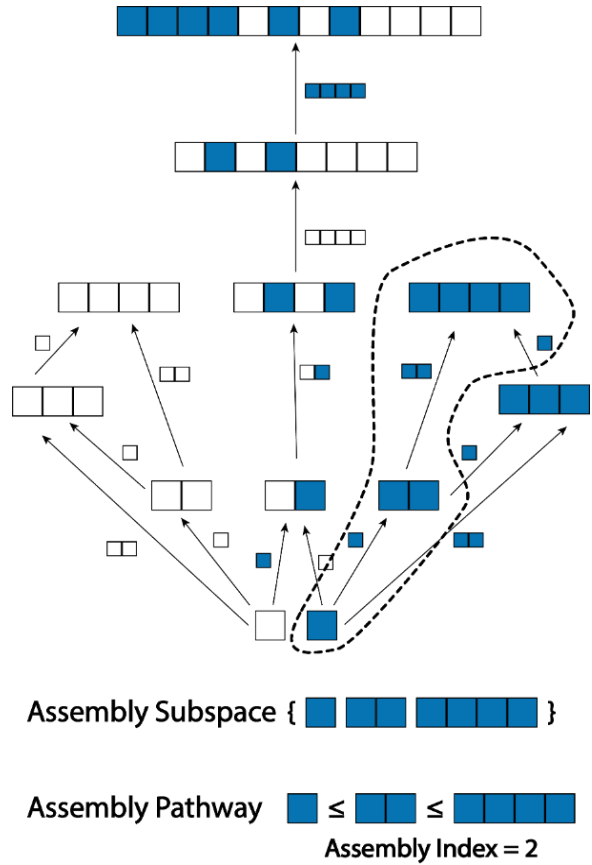

**Figure 3:** An assembly space comprised of objects formed by joining together white and blue blocks. Some of the morphisms have been omitted for clarity. The dotted region is an assembly subspace, and topological ordering of the objects in the subspace represents a minimal assembly pathway for any subspace containing the sequence of four blue boxes.

We define the cardinality and augmented cardinality as the number of objects in the assembly space, where the augmented cardinality excludes the basic objects (defined separately, as this measure is used in the assembly index). We then define an assembly pathway and the assembly index. An assembly pathway is a set of all the objects in an assembly space Γ in some order that respects the reachability relationship ≤, i.e. a topological order. If we take all the rooted assembly subspaces of Γ that contain some object $x$, we then define the assembly index as the augmented cardinality of the smallest rooted assembly subspace that contains $x$. The subspace must be rooted, as otherwise a subspace containing only $x$ would meet this criterion. We use the augmented cardinality of this subspace, as defined above, as defining the assembly index without including basic objects in accord with the physical interpretations that motivated this measure; however, the cardinality could instead be used if desired, and the difference in the measures for any structures with shared basic objects would be require a constant. The assembly index then represents the minimum number of joining operations required to construct object $x$, as illustrated in Figure 2. For a formal definition, see Definition 19 in the SI.

When mapping from assembly space Γ to assembly space Δ through an assembly map f, the assembly index of a mapped object in Δ acts as a lower bound for the assembly index of the original object in Γ. This can allow us, for example, to map an assembly space to another in which finding the assembly index is less computationally intensive in order to calculate a useful lower bound, see Theorem 3 in the SI. The assembly index of an object in any rooted assembly subspace of Γ is an upper bound for the assembly index of the object in Γ, see Lemma 6 in the SI. A split-branched space is an assembly space Γ where for each pair of objects $x, y$ in Γ, $x \leq y$ or $y \leq x$ whenever $V(x \downarrow) \cap V(y \downarrow) \neq \emptyset$ (see Definition 14 in the SI). This means that, other than basic objects, when combining two different objects neither of them can have an assembly pathway that uses objects created in the construction of the other. They may use objects that are considered identical (e.g. the same string) but these are separate objects within the space. Since we can define an assembly map that maps these identical objects to a new space where they map to the same object, the split-branched assembly index for a system is an upper bound for the assembly index on that system.

We use the space of integers under addition to explore these assembly maps, where an addition chain for an integer is a sequence of integers, starting with 1, with each integer in the sequence being the sum of two previous integers, see Figure 4. A minimal addition chain for an integer is the shortest addition chain that terminates in that integer, and the size of that addition chain is equivalent to the pathway assembly index of the integer (after subtracting 1 to account for the single basic object). The objects in this space can be considered as abstract integers, or as representing the size of objects in some other assembly space. See the "Example Applications" section below for more information on addition chains. We model the assembly process as a weighted decision tree where at each level there is a choice of objects that can be formed. The number of choices at each level of the tree is constrained by the number of integers that have the assembly index associated with that level. To obtain the assembly indices, we used data for all minimal addition chain lengths for integers up to 100,000, as published in the Online Encyclopaedia of Integer Sequences (21).

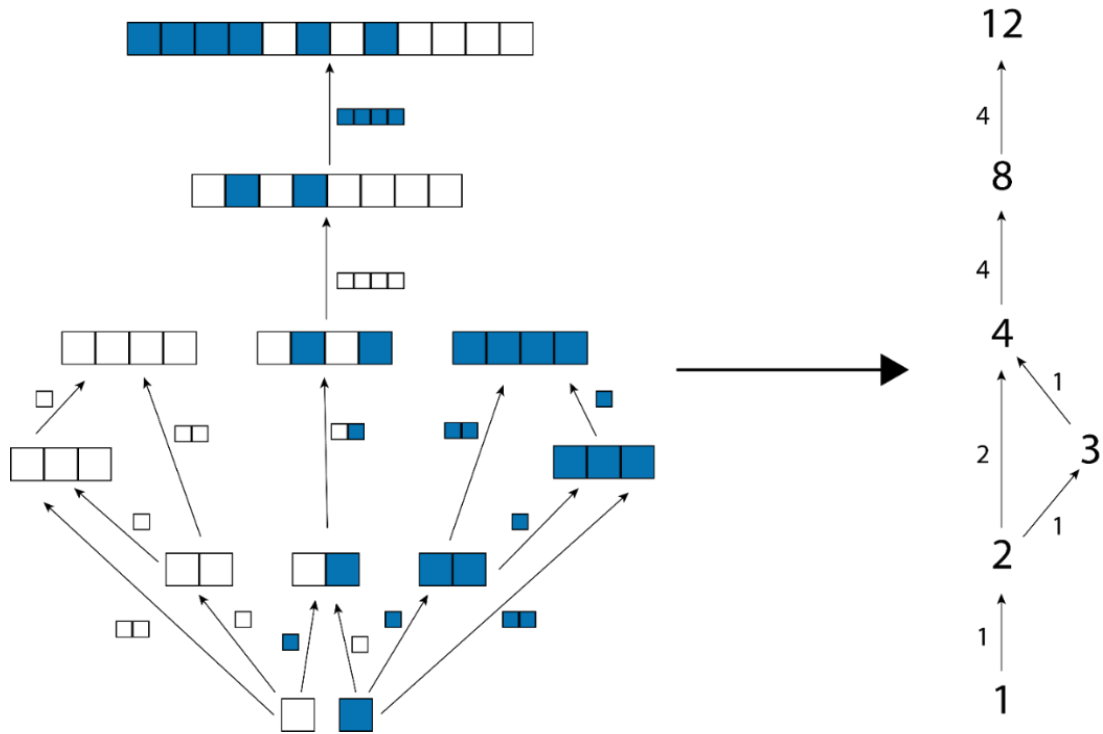

**Figure 4:** An assembly map that maps an assembly space of white and blue blocks onto integers representing the object size.

In the initial case of zero bias, the probability of each step was drawn from a uniform random distribution. In subsequent steps, a value $x$ was drawn from a uniform random distribution between 0 and a value $h$, and the probability of the step was assigned a value $10^x$, subsequently normalised so that all probabilities sum to 1. As $h$ increases, so does the bias of the distribution, with each increase of $x$ by 1 representing a 10-fold increase in likelihood of that choice. We then calculated the probability of the most likely pathway to assess the impact of the bias. In the case of zero bias, at assembly index 25, the integer generated along the most probable pathway will be found has approximately $10^{-7}$ probability of being formed. Increasing the bias to the maximum level $h = 5$, the integer generated along the most probable pathway at assembly index 25 will appear approximately 12% of the time, see Figure 5.

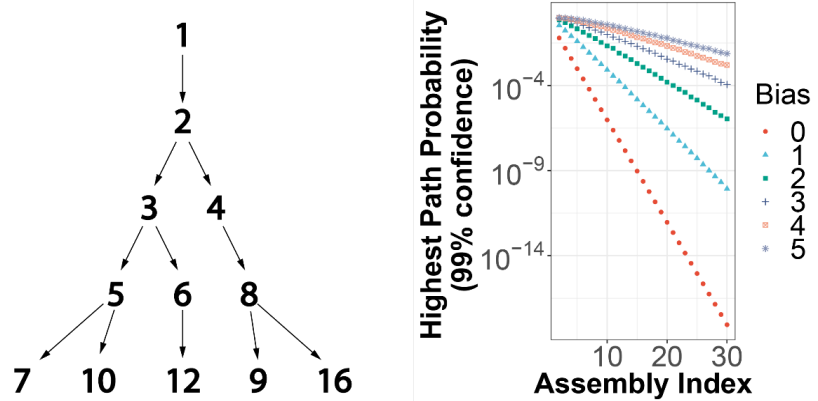

**Figure 5:** Left – Minimal addition chains modelled as a decision tree. Right – probability of the most likely pathway at different levels of bias.

These probabilities will reduce further when considering greater number of choices, such as in situations of higher dimensionality, like strings, grid structures, and graphs (see "Example Applications" section below). In the maximum bias case explored here, where $h = 5$, the choices with $x = 5$ will be 10,000 times more likely than those with $x = 1$. This argument demonstrates that using a pathway assembly model will result in a threshold above which it is unlikely that any specific object would be found, with the threshold depending on the system of objects and joining operations, and the physical limits of the inherent biases present in the process. Even in a significantly biased system, such a threshold will exist, and any objects found in abundance with PA above the threshold will require some process inducing specificity outside of the random (bias) model to form. We consider these additional processes to be biological. Exploration of the processes and biases of a specific system can then be used, along with experimental data, to determine this threshold.

In addition to using the assembly index to determine this biological threshold, it is useful to consider an information measure based on the number of possible structures that can be created using assembly pathways, see Figure 6. One way to do this is to consider a bounded set of possible structures $N$, and then the subset of possible structures with a specified pathway assembly index, $N_{PA}$. The "pathway information" is the amount of uncertainty the pathway assembly index reduces beyond what knowledge of only the size or general composition can provide. In this case, using the approach of Shannon Information (14), the information provided by the Pathway Assembly index, $I_{PA}$, is given by:

$$I_{PA} = \log \frac{|N|}{|N_{PA}|}$$

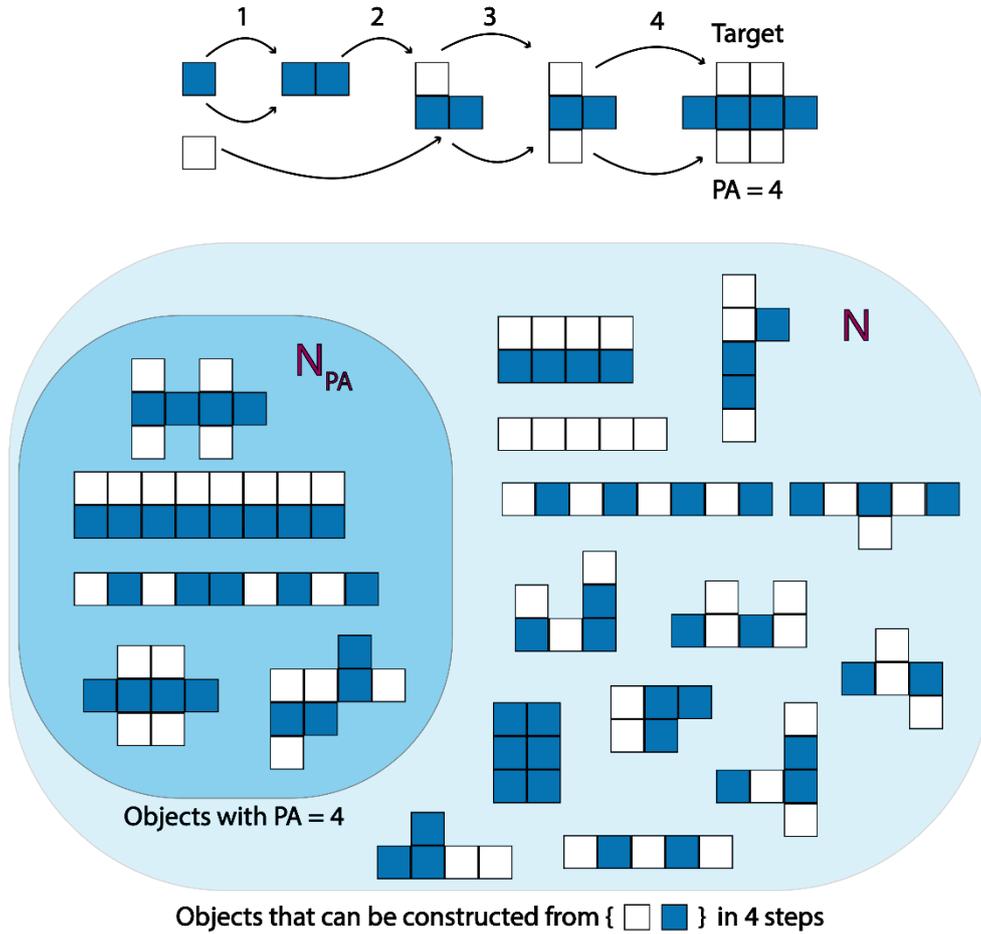

**Figure 6:** Pathway Information. The precise definitions of $N$ and $N_{PA}$ will depend on the specific implementation.

It should be noted that this information measure provides a way of formalizing information over states (size, composition) and over paths (PA) within a common mathematical framework. To calculate $I_{PA}$, there are several possible choices for $N$, all of which must be finite. In one option, for an end product with $PA = x$, $N$ is the set of objects possible that can be created from the same irreducible parts within $x$ steps regardless of PA, and $N_{PA}$ is the subset of those objects with the precise pathway assembly index $PA = x$. This then gives us a measure of the information provided by learning the assembly index, within the context of all objects that could be created by traversing that distance in the assembly space. The difference in utility between the Pathway Assembly index and the pathway information, is that the Assembly Index provides a simple threshold based on pathway length, whereas pathway information can provide an intuition on what the assembly index tells us about the space of possible objects and how much additional information is provided by knowing the paths. The information increases rapidly with Assembly Index, as the space of objects accessible within a given number of steps grows rapidly with the number of steps, see Figure 7. Pseudo code describing the algorithm to calculate the pathway assembly of a given object is described in the SI.

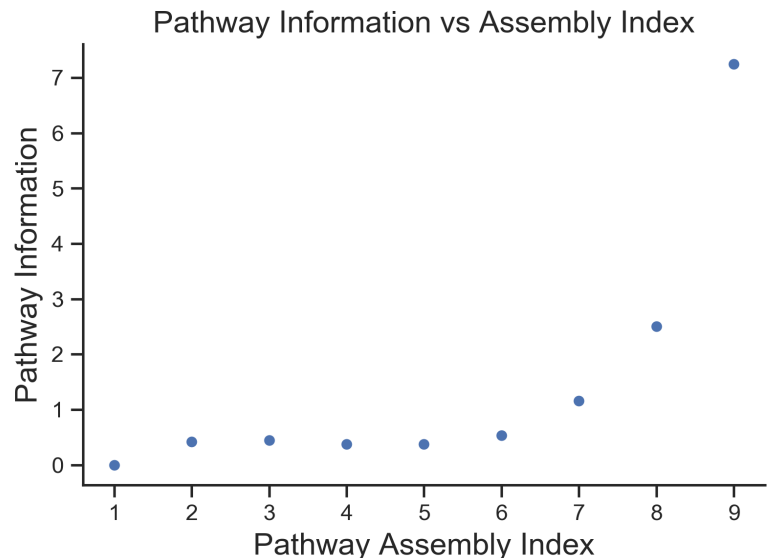

**Figure 7**: Pathway Information vs Pathway Assembly index for strings consisting of 6 x letter "A" and 6 x letter "B". If we have a string with Assembly Index of 9, i.e. it can be constructed in 9 steps, the pathway information is much higher than with Assembly Index 8, as the number of objects that can be constructed in $x$ steps grows much more rapidly than the number of objects with $PA = x$.

**Example Applications**

In the following sections we describe how the pathway assembly approach can be applied to systems of varying dimensionality, see Figure 8.

An addition chain is defined (22) as "a finite sequence of positive integers $1 = a_0 \leq a_1 \ldots \leq a_r = n$ with the property that for all $i > 0$ there exists $j, k$ with $a_i = a_j + a_k$ and $r \geq i > j \geq k \geq 0$. An optimal addition chain is one with the shortest possible length. An example of an optimum addition chain is for $n = 123$ is

$$\{1, 2, 3, 5, 10, 15, 30, 60, 63, 123\}$$

An assembly space $(\Gamma, \Phi)$ for addition chains can be defined where $V(\Gamma) = \mathbb{N}\backslash\{0\} = \{1, 2, 3, ...\}$, the set of positive integers, and for an edge $e \sim [zx]$, $\phi(e) = y$ if and only if $x + y = z$. In this space, an assembly pathway on a subspace representing the assembly index of an integer will be equivalent to an optimum addition chain (subtracting 1 to account for the single basic object). Addition chains can provide a useful lower bound for the assembly index in other assembly spaces, as we can define an assembly map in an assembly space that maps each object to an integer representing the number of basic objects within it (see Figure 3). Addition chains can be generalised to vectorial addition chains (23), in which we define a vectorial addition chain for an k-dimensional vector of natural numbers $n \in \mathbb{N}^k/\{0\}$ (excluding the **0** vector) as a sequence of $a_i \in \mathbb{N}^k/\{0\}$ such that for $-k + 1 \leq i \leq 0$, $a_i$ are the standard basis of unit vectors $\{(1, 0, \ldots, 0), (0, 1, \ldots, 0), \ldots, (0, 0, \ldots, 1)\}$, and for each $i > 0$ there exists $j, k$ with $a_i = a_j + a_k$ and $i > j \geq k$. An example of a vectorial addition chain for [8,8,10] is

$$[[1,0,0], [0,1,0], [0,0,1], [1,1,0], [1,1,1], [2,2,2], [4,4,4], [8,8,8], [8,8,9], [8,8,10]]$$

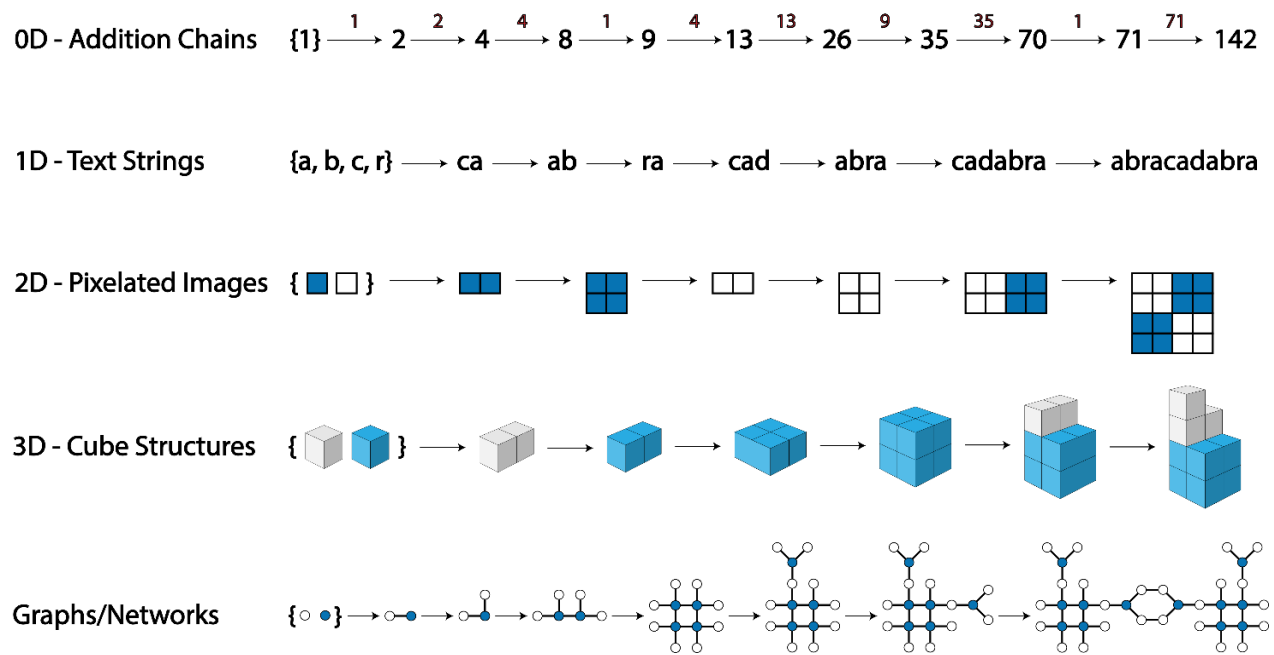

Figure 8: Example assembly pathways for systems of varying dimensionality

We can also define an assembly map from other assembly spaces to vectorial addition chains, where each element in a vector represents the count of a type of basic object (e.g. [1, 2, 3] for 1 red block, 2 blue blocks, 3 green blocks), so this can provide another lower bound. In this case, there also exists a trivial assembly map from vectorial addition chains to addition chains, by summing the vector, so the assembly index on addition chains is a lower bound for the assembly index on vectorial addition chains. In one-dimensional strings we can define an assembly space $(\Gamma, \phi)$ of strings, where each $s \in V(\Gamma)$ is a string and if a string $z$ can be produced by concatenating strings $x$ and $y$, then there exists an edge $e \sim [zx]$ with $\phi(e) = y$, if $z$ can be produced by concatenating $x$ and $y$. There are multiple systems that have string representations, including text strings, binary signals and polymers.

*AAAAAAAAAAAAAAAA:* *{A, AA, AAAA, AAAAAAAA, AAAAAAAAAAAAAAAA}*

*XXBANANAXANANAXX:* *{X, B, A, N, XX, AN, ANAN, ANANA, BANANA, XANANA, BANANAXANANA, XXBANANAXANANA, XXBANANAXANANAXX}*

*GRZXXGZRBVNMMNBV:{G, R, Z, X, B, V, N, M, GR, GRZ, GRZX, GRZXX, GRZXXG, GRZXXGZ, GRZXXGZ, GRZXXGZR, GRZXXGZRB, GRZXXGZRBV, GRZXXGZRBVN, GRZXXGZRBVNM, GRZXXGZRBVNMM, GRZXXGZRBVNMMN, GRZXXGZRBVNMMNB, GRZXXGZRBVNMMNBV}*

**Figure 9:** Examples of text assembly pathways for 16-character strings. The first example demonstrates the shortest possible assembly index of any such string. The second example has a nontrivial assembly pathway, while the third example is a string without any shorter pathway than adding one character at a time. This model assumes that text fragments cannot be reversed when concatenating.

Two other methods for analysing the complexity / information content of strings are the Shannon Information (14) and the Kolmogorov Complexity (15). The Shannon information content of a string is based on the probability of occurrence of its characters. For example, for a string "ABBCCCDDDD", the Shannon entropy (using base 2) is given by

$$H = -\sum_{i \in \{A,B,C,D\}} p_i p_i = -(0.10.1 + 0.20.2 + 0.30.3 + 0.40.4) \simeq 1.84$$

In the case of strings, the probabilities $p_i$ for each character $i$ represent the likelihood of finding it in the string, i.e. the inverse of the count of that character within the string. The Shannon information content is defined as the reduction in entropy (uncertainty) on being presented with some information, and in the case where we are presented with the string itself (reducing entropy to zero) the entropy and information are numerically equal. Unlike Pathway Assembly, Shannon information in this implementation does not consider the structure of the string, e.g. the information content will be the same for "ABBCCCDDDD" as for "ABCDBCDCDD". The Kolmogorov Complexity (15) of an object is the length of the shortest program that outputs that object, in a given Turing-complete language. Although Kolmogorov Complexity is dependent on the language used, it can be shown that the Kolmogorov complexity $C$ in any language $\phi$ can be related to the Kolmogorov complexity in a universal language $U$ by $C_U(x) \leq C\phi(x) + c$ for some constant $c$ (24). If a string cannot be expressed in a universal language by a program shorter than its length, it is considered random. It has been shown that the Kolmogorov complexity is not computable, whereas the Pathway Assembly index is computable (see Theorem 4 in the SI).

We can extend Pathway Assembly to two dimensions by considering a grid of pixels, or coloured boxes, for example a digital image. For simplicity we will consider images with black and white basic objects, although this could be simply extended to greyscale images or colour images (e.g. greyscale images could have 256 basic objects representing different pixel intensities, as in an 8-bit greyscale image). We can define an assembly space with assemblages of black and white pixels as objects. In this space, two assemblages $a$ and $x$ are connected by an edge $e \sim [xa]$ if $a$ is a substructure of $x$. The edge $e$ is labelled as $\phi(e) = b$ with $b$ the complement of $a$ in $x$. In other words, you can connect $a$ and $b$ together to get $x$. A choice can be made about whether to enforce the preservation of orientation, or whether to consider substructures rotated by 90 degrees to be equivalent, and the latter choice can be related to the former by way of an assembly map. An illustration of an assembly pathway in this space can be seen in Figure 10.

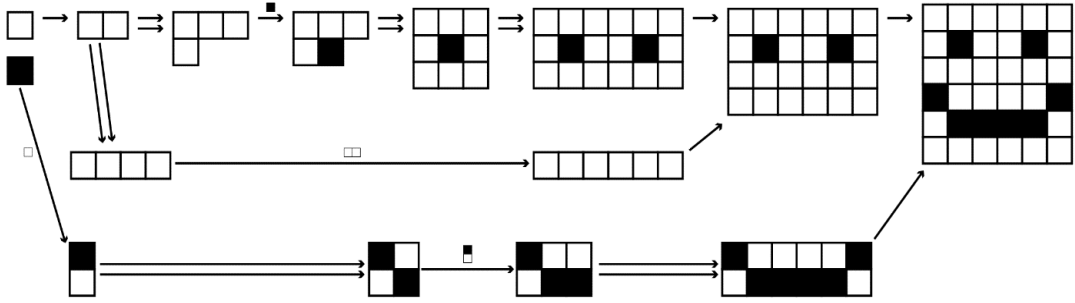

**Figure 10:** Illustrative assembly pathway of a two-dimensional image. This does not necessarily represent the minimal assembly pathway for this shape. Here, images that are rotated are considered equivalent.

The assembly index of an image can be bounded by an assembly map to the one-dimensional case, for example by mapping to a numeric list containing a count of the number of black pixels in each column. It can also be mapped to the space of addition chains as normal and to a reduced representation of the image such as those generated by pooling operations used in convolutional neural networks, or quantisation matrices used in jpeg compression. To extend pathway assembly to three dimensions we can consider structures created out of cubic building blocks as a natural extension of the two-dimensional model. Pathway assembly does not need to be applied to objects as a whole, but can be applied to shared motifs or networks found within the objects (13), which can in some cases map to the problem of cubic building blocks. Pathway assembly, as described here, currently has no simple extension to continuous objects, however we can use an assembly map to define a function that consistently maps similar features to larger block structures, and can calculate the assembly index of that structural motif to explore whether it is over the biological threshold, if found in some abundance. As in the two-dimensional case, the assembly index of cubic structures can be bounded by an assembly map to the two-dimensional case, the one-dimensional case, or to the case of addition chains.

An undirected graph $G(V, E)$ is defined by a set of vertices $V$ and a set of edges $E \subseteq V \times V$. An assembly space for connected graphs (directed or undirected) can be defined where $\Gamma$ is the space of all connected graphs, with the basis set $B$ consisting of a single node. The reachability relationship $\leq$ is defined on $\Gamma$ such that $\phi([G_x, G_a]) = G_b$ if $V_x = V_a \cup V_b$ and $E_x = E_a \cup E_b \cup E_{ab}$ where $E_{ab} \subseteq V_a \times V_b$ and $E_{ab} \neq \emptyset$. In other words, $G_x$ contains all vertices and edges of $G_a$ and $G_b$, and also at least one edge between them. Similar spaces can be defined for graphs that are not necessarily connected by removing the requirement that $E_{ab} \neq \emptyset$. Vertex colours can be incorporated by expanding the basis set $B$. A graph assembly space can also be defined with edges as the basic objects, instead of vertices. Additional constraints allow for the study of spaces of other useful graph structures, for example restriction of vertex degree allows for the study of the space of molecular graphs, which are studied in an upcoming paper. As in the block structures, the assembly space of graphs can be used to analyse objects that have identical network motifs in them while not being identical in other ways. Assembly maps can be defined from the space of graphs to the space of addition chains, as a count of the number of vertices, and also to vectorial addition chains if the vertices are coloured.

There are various other examples where the pathway assembly approach could be used to provide useful analysis of objects. One example is in audio / electromagnetic signals, or music. By utilising notes and silences as basic objects, possibly incorporating frequency/pitch, we could use pathway assembly to distinguish natural signals such as those from a pulsar, or the sound of wind moving through a complex landscape, from sounds such as birdsong or structured communications. In such a system, abundance could be considered to be the same signal from multiple locations, or from the same location but repeated. We can also consider the morphology of apparent geological formations to look for evidence of biological influence in the form of duplicated complex patterns.

Pathway assembly can also be used to define a compression algorithm, similar to the widely known Lempel-Ziv-Welch (LZW) algorithm (25). In the LZW algorithm, repeated portions of text are represented by additional symbols in an expanded character set, and the need for a separate dictionary is removed by building the dictionary in such a way that it can be reconstructed during decompression. In a pathway assembly-based implementation, we could initially calculate an assembly pathway for the string, and then use the additional character set to indicate points at which substrings are duplicated or stored for re-use. It is unlikely that such a compression algorithm would be commercially useful due to the computational complexity of finding a minimal assembly pathway, but analysing compressibility in this way could provide further insights around the information content of string-like objects from an assembly space perspective.

**Conclusions**

The pathway assembly model, and pathway information, can be used to explore the possible ways an object could have formed from its building blocks through random interactions, and we have now built on our initial work (26) by establishing a robust mathematical formalism. By doing so, we can define a threshold above which *extrinsic* information from a biological source would have been required to create an observable abundance of an object because it is too improbable to have formed in abundance otherwise. The pathway assembly of an object, when above the threshold, can be used as an agnostic biosignature, giving a clear indication of the influence of information in constructing objects (e.g. via biological processes) without knowledge of the system that produced the end product. In other words, it can be used to detect biological influence even when we don't know what we are looking for. Of interest is the ability to search for new types of life forms in the lab, alien life on other worlds, as well as identifying the conditions under which the random world embarks on the path towards life, as characterised by the emergence of physical systems that produce objects with high pathway assembly. As such, pathway assembly information might be enable us to not only look for the abiotic to living transition, identifying the emergence of life, but also to identify technosignatures associated with intelligent life with even higher pathway assemblies within a unified quantitative framework. We therefore feel that the concept of pathway assembly can be used to help us explore the universe for structures that must have been produced using an information-driven construction process; in fact we could go as far as to suggest that any such process requiring information is a biological or technological process. This also means that pathway assembly information provides a new window on the problem of understanding the physics of life simply because the physics of information is the physics of life. We believe that such an approach might help us reframe the question from philosophy of what life is (27), to a physics of what life does.

**Acknowledgements**

The authors gratefully acknowledge financial support from the EPSRC (Grant Nos EP/R01308X/1, EP/L023652/1, EP/P00153X/1), the ERC (project 670467 SMART-POM), the John Templeton Foundation Grant ID 60625 and Grant ID 61184 and the National Aeronautics and Space Administration through grant NNX15AL24G S02. We thank Dr. Cole Mathis and Prof. Paul Davies for useful discussions.

**Author Contributions**

L.C. conceived of the overall concept and developed the algorithm together with S.M.M. A.R.G.M explored the initial mathematical description, and this was expanded and validated by D.G.M and S.I.W. L.C. and S. M. M. wrote the manuscript with input from all the authors.

# Supplementary Information: Quantifying the pathways to life using assembly spaces

Stuart M. Marshall,[1] Douglas G. Moore,[2] Alastair R. G. Murray,[1] Sara I. Walker,[2,3*] and Leroy Cronin[1*]

[1] School of Chemistry, University of Glasgow, Glasgow, G12 8QQ, UK.

[2] BEYOND Center for Fundamental Concepts in Science, Arizona State University, Tempe, AZ, USA

[3] School of Earth and Space Exploration, Arizona State University, Tempe, AZ, USA

*Corresponding author email: Lee.Cronin@glasgow.ac.uk, sara.i.walker@asu.edu

## 1. Introduction

This formalism arose as a means of rigorously describing the "simplest" way of assembling a given object by combining basing building blocks. With this in mind, we consider a universe of objects and binary relations between them signifying that one can be compose with *some* third object to create the second. Each relation then is associated with that third object. In this setting, the concrete rules or laws describing this assembly processes are neglected though they are certainly necessary for initially constructing the space. We quickly come to the conclusion that a graph formalization is appropriate when we consider this kind of process in general, but a more general mathematical structure than the standard concept of a graph is necessarily; namely *quivers*[1]. As such, the fundamental mathematical structure we define and explore herein, referred to as an *assembly space*, can be described as an acyclic quiver with edges labeled with vertices in the quiver which satisfies three simple axioms. With assembly spaces defined, we are able to define an *assembly index* for each object in the space which characterizes how directly that object can be assembled. Further, we prove several axioms relating to method of bounding the assembly index for a given object and algorithms for computing or approximating it.

To facilitate this exposition, we have broken the text into three sections. First, in Section 2. we describe the graph-theoretic prerequisites associated with quivers and morphisms (mappings) between them. We proceed then to define assembly spaces, subspaces and maps between them, and to prove several lemmas in Section 3. Finally, we define the assembly index, prove that it is computable and two methods for bounding it above and below, and present algorithms for computing or approximating it in Section 4.

## 2. Graph-Theoretic Prerequisites

We begin by considering a set of objects, possibly infinitely many objects, which can be combined in various ways to produce others. If an object $a$ can be combined with some other object to yield an object $b$, we represent the relationship between $a$ and $b$ by drawing a directed edge or arrow from $a$ to $b$. Altogether, this structure is a quiver, also called a multigraph, as we allow for the possibility that there is more than one way to produce $b$ from $a$; that is, there may be more than one edge from $a$ to $b$.

---

[1] Many texts refer to this structure as a *multigraph*, with the term *quiver* preferred in settings where the edges represent morphism or processes rather than simply relationships. Since we view the "relations" as an active process of combination, we prefer *quiver* in this text. The reader would lose nothing by mentally substituting the terms.

**Definition 1.** A **quiver** $\Gamma$ consists of

1. a set of vertices $V(\Gamma)$
2. a set of edges $E(\Gamma)$
3. a pair of maps $s_\Gamma, t_\Gamma \colon E(\Gamma) \to V(\Gamma)$

For an edge $e \in E(\Gamma)$, $s_\Gamma(e)$ is referred to as the source and $t_\Gamma(e)$ the target of the edge, and we will often leave off the subscripts when the context is clear, e.g. $s$ and $t$. We will often describe an edge $e \in E(\Gamma)$ with $s(e) = a$ and $t(e) = b$ as $e \sim [ba]$. This does not mean that $e$ is a unique edge with endpoints $a$ and $b$; it is possible that two edges $e \neq f$ have the same endpoints $e \sim f \sim [ba]$.

From here, we consider paths, that is sequences of edges, which describe the process of sequentially combining objects to yield intermediate objects and ultimately some terminal object.

**Definition 2.** If $\Gamma$ is a quiver, a **path** $\gamma = a_n \ldots a_1$ in $\Gamma$ of length $n \geq 1$ is a sequence of edges such that $t(a_i) = s(a_{i+1})$ for $1 \leq i \leq n - 1$. The functions $s$ and $t$ can be extended to paths as $s(\gamma) = s(a_1)$ and $t(\gamma) = t(a_n)$. We write $|\gamma|$ to denote the length, or number of edges, in the path. Additionally, for each vertex $x \in \Gamma$ there is a **zero path**, denoted $e_x$, with length 0 and $s(e_x) = t(e_x) = x$.

A natural point is that combining two objects should never yield something that can be used to create either of those objects. Essentially, there are no directed cycles – sequences of edges that form a closed cycle – within the quiver.

**Definition 3.** A path $\gamma$ in a quiver $\Gamma$ is a **directed cycle** if $|\gamma| \geq 1$ with $t(\gamma) = s(\gamma)$.

**Definition 4.** A quiver $\Gamma$ is **acyclic** if it has no directed cycles.

We can think of an object $b$ as being *reachable* from an object $a$ if there's a path from $a$ to $b$, and this relationship forms a partial ordering on the quiver if the quiver is acyclic.

**Definition 5.** Let $\Gamma$ be an acyclic quiver and let $x, y \in V(\Gamma)$. We say $y$ is **reachable** from $x$ if there exists a path $\gamma$ such that $s(\gamma) = x$ and $t(\gamma) = y$, where $|\gamma| \geq 0$.

**Lemma 1.** *Let $\Gamma$ be an acyclic quiver, and define a binary relation $\leq$ on the vertices of $\Gamma$ such that $x \leq y$ if and only if $y$ is reachable from $x$. $(V(\Gamma), \leq)$ is a partially ordered set, and $\leq$ is referred to as the **reachability relation** on $\Gamma$.*

*Proof.* For $\leq$ to be a partial ordering on $V(\Gamma)$, we need to show that it is reflexive, transitive and antisymmetric. Reflexivity follows directly from the definition of reachability as $x$ is reachable from itself via the zero path $e_x$. To show transitivity, let $a \leq b$ and $b \leq c$. If $a = b$ or $b = c$, then we're done. Otherwise there are paths $\gamma_{ba} = u_m \ldots u_1$ from $a$ to $b$ and $\gamma_{cb} = v_n \ldots v_1$ from $b$ to $c$. The composite path $\gamma_{cb} \circ \gamma_{ba} = v_n \ldots v_1 u_n \ldots u_1$ is a path from $a$ to $c$; thus $c$ is reachable from $a$ so that $a \leq c$. Now consider antisymmetry and suppose that $a \leq b$ and $b \leq a$. Then there exist paths $\gamma_{ba}$ and $\gamma_{ab}$ from $a$ to $b$ and $b$ to $a$, respectively. Then $\gamma_{ab} \circ \gamma_{ba}$ is a path from $a$ to itself. Since $\Gamma$ is acyclic, this implies that $\gamma_{ab} \circ \gamma_{ba} = e_a$, and consequently that $\gamma_{ab} = \gamma_{ba} = e_a$. Thus $a = b$ and $\leq$ is antisymmetric.

∎

The idea of reachability allows us to think of all objects that are reachable from (or above) a given object $x$, the upper quiver of $x$. Similarly, we can think of all objects that can reach $x$, the lower quiver.

**Definition 6.** Let $\Gamma$ be an acyclic quiver and let $\leq$ be the reachability relation on it. The **upper quiver** of $x \in V(\Gamma)$ is $x \uparrow$ with vertices $V(x \uparrow) = \{y \in V(\Gamma) \mid x \leq y\}$, edges $E(x \uparrow) = \{e \in E(\Gamma) \mid s_\Gamma(e), t_\Gamma(e) \in V(x \uparrow)\}$, $s_{x\uparrow} = s_\Gamma|_{E(x\uparrow)}$, and $t_{x\uparrow} = t_\Gamma|_{E(x\uparrow)}$. The **lower quiver** of $x \in V(\Gamma)$ is $x \downarrow$ with vertices $V(x \downarrow) = \{y \in V(\Gamma) \mid y \leq x\}$, edges $E(x \downarrow) = \{e \in E(\Gamma) \mid s_\Gamma(e), t_\Gamma(e) \in V(x \downarrow)\}$, $s_{x\downarrow} = s_\Gamma|_{E(x\downarrow)}$, and $t_{x\downarrow} = t_\Gamma|_{E(x\downarrow)}$.

Similarly, the upper quiver of a subset $Q \subseteq V(\Gamma)$ in $\Gamma$ is $Q \uparrow$ with vertices $V(Q \uparrow) = \{y \in V(\Gamma) \mid (\exists q \in Q) q \leq y\}$, edges $E(Q \uparrow) = \{e \in E(\Gamma) \mid s_\Gamma(e), t_\Gamma(e) \in V(Q \uparrow)\}$, $s_{Q\uparrow} = s_\Gamma|_{E(Q\uparrow)}$, and $t_{Q\uparrow} = t_\Gamma|_{E(Q\uparrow)}$. The lower set of a subset is defined dually.

Going further, we can consider those objects that cannot be reached as *minimal* and those that cannot reach anything as *maximal*. An object which can be reached by finitely many objects is called *finite*.

**Definition 7.** Let $\Gamma$ be an acyclic quiver, $\leq$ be the reachability relation on it and $x$ a vertex in $\Gamma$. Then $x$ is said to be **maximal** in $\Gamma$ if, whenever $x \leq y$ in $\Gamma$, we have $x = y$. Dually, $x$ is **minimal** in $\Gamma$ if, whenever $y \leq x$ in $\Gamma$, we have $x = y$. The set of all maximal vertices of $\Gamma$ is denoted $\max(\Gamma)$ with $\min(\Gamma)$ defined dually.

**Definition 8.** A quiver $\Gamma$ is said to be **finite** if its vertex and edge sets are both finite. Similarly, a vertex $x$ in a quiver $\Gamma$ is said to be finite if $x \downarrow$ in $\Gamma$ is a finite quiver.

With this idea of a quiver of objects defined, we can consider asking about subsets of objects and relations between them in the context of the quiver as a whole.

**Definition 9.** Let $\Gamma$ and $\Gamma'$ be quivers. Then $\Gamma'$ is a **subquiver** of $\Gamma$ if $V(\Gamma') \subseteq V(\Gamma)$, $E(\Gamma') \subseteq E(\Gamma)$, $s_{\Gamma'} = s_\Gamma|_{E(\Gamma')}$ and $t_{\Gamma'} = t_\Gamma|_{E(\Gamma')}$. We will denote this relationship as $\Gamma' \subseteq \Gamma$.

**Lemma 2.** *If $X$, $Y$ and $Z$ are quivers such that $X \subseteq Y$ and $Y \subseteq Z$, then $X \subseteq Z$. That is, the binary relation $\subseteq$ on quivers is transitive.*

*Proof.* Suppose $X$, $Y$ and $Z$ are quivers with $X \subseteq Y$ and $Y \subseteq Z$. Then $V(X) \subseteq V(Y) \subseteq V(Z)$, so that $V(X) \subseteq V(Z)$. Similarly, $E(X) \subseteq E(Z)$. Next, since $s_X = s_Y|_{E(X)}$, $s_Y = s_Z|_{E(Y)}$ and

$E(X) \subseteq E(Y)$, $s_X = s_Z|_{E(X)}$. The same argument applies to show that $t_X = t_Z|_{E(X)}$. Thus $X \subseteq Z$, so that $\subseteq$ is transitive.

∎

Finally, we will need to consider how to map one quiver to another in a consistent fashion, maintaining the basic relational structure of the original quiver.

**Definition 10.** Let $\Gamma$ and $\Gamma'$ be quivers. A **quiver morphism**, denoted $m: \Gamma \to \Gamma'$, consists of a pair $m = (m_v, m_e)$ of functions $m_v: V(\Gamma) \to V(\Gamma')$ and $m_e: E(\Gamma) \to E(\Gamma')$ such that $m_v \circ s_\Gamma = s_{\Gamma'} \circ m_e$ and $m_v \circ t_\Gamma = t_{\Gamma'} \circ m_e$. That is, the following diagrams commute:

$$\begin{array}{ccc} E(\Gamma) & \xrightarrow{m_e} & E(\Gamma') \\ \downarrow{s_\Gamma} & & \downarrow{s_{\Gamma'}} \\ V(\Gamma) & \xrightarrow{m_v} & V(\Gamma') \end{array} \qquad \begin{array}{ccc} E(\Gamma) & \xrightarrow{m_e} & E(\Gamma') \\ \downarrow{t_\Gamma} & & \downarrow{t_{\Gamma'}} \\ V(\Gamma) & \xrightarrow{m_v} & V(\Gamma'). \end{array}$$

## 3. Assembly Spaces, Subspaces and Maps

Up to this point, we have focused on the binary idea that one object can be used as a structural ingredient of another. However, we need something more if we want to capture the idea that two things must be combined in order to assemble another. We do this by labeling the edges of the quiver with the object that the source of the edge is combined with to produce the target. Further, this labeling has to be consistent. If $a$ can be combined with $b$ to yield $c$, then $b$ can be combined with $a$ to yield $c$. Additionally, we require that there exists a set of minimal objects – building blocks from which all other objects can be assembled.

**Definition 11.** An **assembly space** is an acyclic quiver $\Gamma$ together with an edge-labeling map $\phi: E(\Gamma) \to V(\Gamma)$ which satisfies the following axioms:

4. $\min(\Gamma)$ is finite and non-empty

5. $\Gamma = \min(\Gamma) \uparrow$

6. If $a$ is an edge from $x$ to $z$ in $\Gamma$ with $\phi(a) = y$, then there exists an edge $b$ from $y$ to $z$ with $\phi(b) = x$.

Such an assembly space is denoted $(\Gamma, \phi)$ or simply $\Gamma$ where appropriate. We will continue to write $x \in \Gamma$ to mean that $x$ is a vertex of the quiver.

**Definition 12.** The set of minimal vertices of an assembly space $\Gamma$ is referred to as the **basis** of $\Gamma$ and is denoted $B_\Gamma$. Elements of the basis are referred to as basic objects, basic vertices or basic elements.

**Definition 13.** An **assembly pathway** of an assembly space $\Gamma$ is any topological ordering of the vertices of $\Gamma$ with respect to the reachability relation.

**Definition 14.** An assembly space $\Gamma$ with reachability relation $\leq$ is said to be **split-branched** if for all $x, y \in \Gamma$, $x \leq y$ or $y \leq x$ whenever $V(x \downarrow) \cap V(y \downarrow) \neq \emptyset$.

**Definition 15.** Let $(\Gamma, \phi)$ and $(\Gamma', \psi)$ be assembly spaces. Then $(\Gamma', \psi)$ is an **assembly subspace** of $(\Gamma, \phi)$ if $\Gamma'$ is a subquiver of $\Gamma$ and $\psi = \phi|_{E(\Gamma')}$. This relationship is denoted as $(\Gamma, \phi) \subseteq (\Gamma', \psi)$, or simply $\Gamma \subseteq \Gamma'$ when there is no ambiguity.

**Definition 16.** Let $\Gamma'$ be an assembly subspace of $\Gamma$. Then $\Gamma'$ is rooted in $\Gamma$ if $B_{\Gamma'} \subseteq B_\Gamma$ as sets.

**Lemma 3.** *Let $U$, $V$ and $W$ be assembly spaces with $U \subseteq V$ and $V \subseteq W$, then $U \subseteq W$. Further, if $U$ is rooted in $V$, and $V$ is rooted in $W$, then $U$ is rooted in $W$.*

*Proof.* Let $(U, \phi_U)$, $(V, \phi_V)$ and $(W, \phi_W)$ be assembly spaces such that $(U, \phi_U) \subseteq (V, \phi_V)$ and $(V, \phi_V) \subseteq (W, \phi_W)$. Since $U, V$ and $W$ are quivers, $U \subseteq W$ by the transitivity of $\subseteq$ on quivers. Further, since $\phi_U = \phi_V|_{E(U)}$, $\phi_V = \phi_W|_{E(V)}$ and $E(U) \subseteq E(W)$, we have $\phi_U = \phi_W|_{E(W)}$. Thus, $(U, \phi_U) \subseteq (W, \phi_W)$. That is, $\subseteq$ is transitive on assembly spaces. If $U$ is rooted in $V$ and $V$ is rooted in $W$, then $B_U \subseteq B_V \subseteq B_W$. That is, $U$ is rooted in $W$.

∎

**Lemma 4.** *Let $(\Gamma, \phi)$ be an assembly space and let $x \in \Gamma$. If $e \sim [ba]$ is an edge in $\Gamma$ with $a, b \in x \downarrow$, then $\phi(e) \in x \downarrow$.*

*Proof.* Since $\Gamma$ is an assembly space, we have $\phi(e) \leq b$ where $\leq$ is the reachability relation on $\Gamma$. By construction, $b \leq x$ so that $\phi(e) \leq x$. Therefore $\phi(e) \in x \downarrow$.

∎

**Lemma 5.** *Let $(\Gamma, \phi)$ be an assembly space and let $x \in \Gamma$. Then $(x \downarrow, \phi|_{x\downarrow})$ is a rooted assembly subspace of $\Gamma$.*

*Proof.* We first show that $(x \downarrow, \phi|_{x\downarrow})$ is an assembly space. Since $(\Gamma, \phi)$ is an assembly space, it is the upper set of its basis $B_\Gamma$. As such $\min(x \downarrow)$ is a non-empty subset of $B_\Gamma$ and $x \downarrow = \min(x \downarrow) \uparrow$ giving us axiom 1. The remaining axiom follows directly from lemma 4. What's more, we already have that $\min(x \downarrow) = B_{x\downarrow} \subseteq B_\Gamma$, so $x \downarrow$ is rooted in $\Gamma$.

∎

**Definition 17.** Let $(\Gamma, \phi)$ and $(\Delta, \psi)$ be assembly spaces. An assembly map is a quiver morphism $f: \Gamma \to \Delta$ such that $\psi \circ f_e = f_v \circ \phi$. That is, the following diagram commutes:

$$\begin{array}{ccc} E(\Gamma) & \xrightarrow{f_e} & E(\Delta) \\ \downarrow \phi & & \downarrow \psi \\ V(\Gamma) & \xrightarrow{f_v} & V(\Delta). \end{array}$$

Our first theorem provides a basis for the lower bounds developed in the next section. The essential point is that the image of an assembly space under an assembly space is an assembly space.

**Theorem 1.** *If $f: \Gamma \to \Delta$ is an assembly map between assembly spaces $(\Gamma, \phi)$ and $(\Delta, \psi)$, then $(f(\Gamma), \varphi)$ with $\varphi = \psi|_{E(f(\Gamma))}$ is an assembly subspace of $\Delta$.*

*Proof.* Since $f$ is a quiver morphism and $\Delta$ is acyclic, $f(\Gamma)$ is an acyclic subquiver of $\Delta$. By construction, $\varphi = \psi|_{E(f(\Gamma))}$. What remains is to prove the three assembly space axioms. Let $f_v$ and $f_e$ be the vertex and edge maps comprising $f$.

> **Axiom 1** We must show that $\mathbf{min}(f(\Gamma))$ is finite and non-empty.
> We start by showing that $\min(f(\Gamma)) \neq \emptyset$. To see this, consider an element $b \in f_v(\min(\Gamma))$, and suppose there exists a path from an element $x \in f(\Gamma)$ to $b$. Then let $v \in f_v^{-1}(x)$ and let $\gamma$ be a path from a basic element $u \in \min(\Gamma)$ to $v$ – which must exist since $\Gamma$ is an assembly space. The image of this path is a path in $f(\Gamma)$ from $f_v(u)$ to $x$, and consequently a path from $f_v(u)$ to $b$. Since $f_v(\min(\Gamma))$ is finite, we can repeat this process beginning with the newly identified element of $f_v(\min(\Gamma))$ only finitely many times before a cycle is formed. However, that cycle must have length zero since $\Delta$ contains no cycles of greater length. As such, the final element of $f_v(u)$ produced is in $\min(f(\Gamma))$ since there is nothing below it in $f(\Gamma)$. Thus, $\min(f(\Gamma))$ is non-empty.
>
> We now show that, $\min(f(\Gamma))$ is in fact finite. In particular, $\min(f(\Gamma))$ is a subset of a finite set, namely $f_v(\min(\Gamma))$, so it too is finite. Let $x \in \min(f(\Gamma))$. Then there exists an element $b \in f_v^{-1}(x)$ and at least one path $\gamma$ from a basic element $a \in \min(\Gamma)$ to $b$. The image of $\gamma$ under $f$ is a path in $f(\Gamma)$ from $f_v(b)$ to $x$. Since $x$ is minimal in $f(\Gamma)$, the only paths that terminate at $x$ are zero paths. Thus $f_v(b) = x$. Since $x$ was a generic element of $\min(f(\Gamma))$, every element of $\min(f(\Gamma))$ is the image of a basic element of $\Gamma$. That is $\min(f(\Gamma)) \subseteq f_v(\min(\Gamma))$, so it's finite.

**Axiom 2** Next we prove that $f(\Gamma) = \min(f(\Gamma)) \uparrow$. Let $x$ be an element of $f(\Gamma)$. We aim to show that there exists a path from a basic element of $f(\Gamma)$ to $x$. Let $b$ be an element of $\Gamma$ which maps to $x$ under application of $f$. Then, since $\Gamma$ is an assembly space, we know there exists at least one path, $\gamma$ from a basic element of $\Gamma$, say $a \in \min(\Gamma)$, to $b$. The image of this path in $f(\Gamma)$ is itself a path from $f_v(a)$ to $x$, namely $f_e(\gamma)$. If $f_v(a)$ is a basic element of $f(\Gamma)$ the we are done. Otherwise, we can use the processes described in the proof of Axiom 1 to construct a path from $f_v(a)$ through basic and non-basic elements which will ultimately terminate at a basic element. Composing this path with $f_e(\gamma)$ then yields a path from a basic element to $x$. As such, every element of $f(\Gamma)$ is above at least one basic element of $f(\Gamma)$, i.e. $f(\Gamma) = \min(f(\Gamma)) \uparrow$.

**Axiom 3** We now must show that for every edge $a \in E(f(\Gamma))$ with $a \sim [zx]$ and $\varphi(a) = y$, then there exists an edge $b \in E(f(\Gamma))$ with $b \sim [zy]$ and $\varphi(b) = x$. To see that this is the case, take $a$ as described. Then there exists an edge $u \in f_e^{-1}(a)$ in $\Gamma$ with $u \sim [rq]$, $f_v(r) = z$, $f_v(q) = y$ and $\phi(u) = p$. Since $\Gamma$ is an assembly space, there exists an edge $v \in E(\Gamma)$ with $v \sim [rp]$ and $\phi(v) = q$. The commutivity property of assembly maps then gives us $\varphi(f_e(v)) = f_v(\phi(v)) = f_v(q) = y$. Calling $f_v(p) = x$ we then have an edge in $f(\Gamma)$, namely $f_e(v)$, which terminates at $z$ and is labeled as $y$. This satisfies Axiom 3.

∎

## 7. The Assembly Index

This final section turns to the definition and computation of the assembly index, a measure of how directly an object can be constructed from basic objects.

**Definition 18.** The **cardinality** of an assembly space $(\Gamma, \phi)$ is the cardinality of the underlying quiver's vertex set, $|V(\Gamma)|$. The augmented cardinality of the $(\Gamma, \phi)$ with basis $B_\Gamma$ is $|V(\Gamma) \backslash B_\Gamma| = |V(\Gamma)| - |B_\Gamma|$.

**Definition 19.** The **assembly index** $c_\Gamma(x)$ of a *finite* object $x \in \Gamma$ is the minimal augmented cardinality of all rooted assembly subspaces containing $x$. This can be written $c(x)$ when the relevant assembly space $\Gamma$ is clear from context.

## 4.1 Bounds on the Assembly Index

First of all, we can bound the assembly index of an object from above by computing the assembly index of that object in a rooted subspace. Essentially, since assembly subspace generally has fewer edges, there are fewer "shortcuts" to assembly the giving object.

**Lemma 6.** *Let X be an assembly space and Y a rooted assembly subspace of X. For every finite $y \in Y$, the assembly index of y in Y is greater than or equal to the assembly index of y in X. That is, $c_Y(y) \geq c_X(y)$ for all $y \in Y$.*

*Proof.* Let $y \in Y$ and suppose $c_Y(y) < c_X(y)$. Then there exists a rooted assembly subspace $Z \subseteq Y$ containing $y$ such that $|Z \backslash B_Z| = c_Y(y)$. But, by the transitivity of subset inclusion (lemma 3) $Z$ is a rooted assembly subspace of $X$. But if that's the case, there exists a rooted assembly subspace of $X$ with augmented cardinality less than $c_X(y)$, namely $Z$; a contradiction.

∎

Since the lower quiver of an object $x$ is a rooted assembly subspace, we know the assembly index of the object in $x \downarrow$ bounds the real assembly index of the object from above. However, we can do better – $c_\Gamma(x) = c_{x\downarrow}(x)$. This is a very powerful result as it allows us any computational approaches aiming to compute $c(x)$ to focus only on the objects below $x$.

**Theorem 2.** *Let $\Gamma$ be an assembly space and let $x \in \Gamma$ be finite. Then $c_\Gamma(x) = c_{x\downarrow}(x)$.*

*Proof.* Since $x \downarrow$ is finite, we need only consider finite, rooted assembly subspaces of $\Gamma$. Let $\Delta \subseteq \Gamma$ be such a subspace containing $x$, and suppose that $\Delta \nsubseteq x \downarrow$. Let $y \in \Delta$ such that $y \notin x \downarrow$, then $(\Delta \backslash y \uparrow)$ is a rooted assembly subspace of $\Gamma$ containing $x$ with augmented cardinality strictly less than $\Delta$. As such $|\Delta \backslash B_\Delta| \neq c_\Gamma(x)$. In other words, if $\Delta$ is not a subspace of $x \downarrow$, then it cannot have the augmented cardinality $c_\Gamma(x)$. Thus, by contrapositive if $|\Delta \backslash B_\Delta| = c_\Gamma(x)$, then $\Delta \subseteq x \downarrow$. Since $\Delta$ is rooted in $\Gamma$, it must also be rooted in $x \downarrow$. Therefore, if a rooted subspace of $\Gamma$ has the minimal augmented cardinality in $\Gamma$, it must be a rooted assembly subspace of $x \downarrow$. This implies that $c_\Gamma(x) \geq c_{x\downarrow}(x)$. Additionally, by lemma 6, $c_\Gamma(x) \leq c_{x\downarrow}(x)$. Then $c_\Gamma(x) = c_{x\downarrow}(x)$.

∎

Finally, assembly maps allow us to place lower-bounds on the assembly index – the assembly index of the image of an object bounds the object's actual assembly index below. In other words, we can place lower bounds on the assembly index of an object by mapping the assembly space into a simpler space and computing the assembly index there.

**Theorem 3.** *If $f: \Gamma \to \Delta$ is an assembly map, then $c_{f(\Gamma)}(f(x)) \leq c_\Gamma(x)$ for all finite $x \in \Gamma$.*

*Proof.* Let $\Sigma \subseteq \Gamma$ be an assembly subspace containing $x$ with $|\Sigma \backslash B_\Sigma| = c_\Gamma(x)$. The restriction of $f$ to $\Sigma$ is an assembly map $f^*: \Sigma \to f(\Gamma)$. Then we have

$$\begin{aligned}
|\Sigma \backslash B_\Sigma| &\geq |f^*(\Sigma \backslash B_\Sigma)| \\
&= |f^*(\Sigma \backslash B_\Sigma) \cap (f^*(\Sigma) \backslash B_{f^*(\Sigma)})| + |f^*(\Sigma \backslash B_\Sigma) \cap B_{f^*(\Sigma)}| \\
&\geq |f^*(\Sigma \backslash B_\Sigma) \cap (f^*(\Sigma) \backslash B_{f^*(\Sigma)})|.
\end{aligned}$$

As an assembly map, $f^*$ maps basis elements of $\Sigma$ onto basis elements of $f^*(\Sigma)$. So every for $u \in f^*(\Sigma) \backslash B_{f^*(\Sigma)}$, there exists a $v \in \Sigma \backslash B_\Sigma$ such that $f^*(v) = u$. This gives us

$$\begin{aligned}
c_\Gamma &= |\Sigma \backslash B_\Sigma| \\
&\geq |f^*(\Sigma \backslash B_\Sigma) \cap (f^*(\Sigma) \backslash B_{f^*(\Sigma)})| \\
&= |f^*(\Sigma) \backslash B_{f^*(\Sigma)}| \\
&\geq c_{f(\Gamma)}(f(x)).
\end{aligned}$$

## 4.2 Computability and Algorithms

**Theorem 4.** If $\Gamma$ is an assembly space and $x \in \Gamma$ is finite, then $c(x)$ is computable.

*Proof.* As shown in the proof of theorem 2, every rooted assembly subspace with minimal augmented cardinality and containing $x$ is a minimal rooted assembly subspace of $x \downarrow$. Since $x$ is finite, $x \downarrow$ is finite, the set of assembly subspace of $x \downarrow$ is finite, and each such subspace is finite. Consequently, the basis of each subspace is computable. As such, the set of all rooted subspaces is computable. The cardinality of each subspace is computable, so the set of cardinalities of all rooted subspaces is computable. Finally, the minimum of a finite set of natural numbers is computable. Therefore, $c_\Gamma(x)$ is computable.

∎

An algorithm for finding the pathway assembly index of an object within an assembly subspace is described below.

The Assembly Index in assembly space $\Gamma \equiv (\Gamma, \phi)$ of a target object $t \in \Gamma$, with basic objects $B \subseteq \Gamma$.

```
Function Main(B, t)
Global Variable PA // the pathway assembly index
Set PA = upper bound of assembly index + |B|
AssemblyIndex(B, t)
Return PA - |B|
End Function
Function AssemblyIndex(S, t)
    For each pair of objects s₁, s₂ ∈ S
        If there exists an edge e~[ts₁] with φ(e) = s₂ and PA > |S ∪ t|
            PA = |S ∪ T|
        Else there exists an edge e~[us₁] with φ(e) = s₂ for some u ∈ Γ
            AssemblyIndex(Γ, S ∪ u, t)
        End If
    End For
End Function
```

We have also defined above the Split-Branched Assembly Index. Calculation of this index can be more computationally tractable than the assembly index, as often a lower number of pathways will need to be enumerated. An algorithm to calculate this index is shown below.

The Split-Branched Assembly Index in assembly space $\Gamma \equiv (\Gamma, \phi)$ of a target object $t \in \Gamma$ with, with basic objects $B \subseteq V(\Gamma)$

```
Function SplitBranchedAssemblyIndex(Γ, B, t, I)
    Set PA = upper bound of assembly index for t
    For each partition of U into connected sub-objects Γ_P = {Γ₁ … Γₙ}
        Set PartitionIndex = 0
        Partition U_P into K = {{Γ₁₁, …, Γ₁ᵢ}, {Γ₂₁, …, Γ₂ⱼ}, …, {Γₘ₁ … Γₘₖ}}
        Where for each Kₙ, the Γₙₓ are identical for all x
        For each Kᵢ ∈ K
            If Kᵢ₁ ∈ B
                PartitionIndex += 1
            Else
                PartitionIndex += SplitBranchedAssemblyIndex(Γ, B, Kᵢ₁)
```

```
                                                       + |K_i| − 1
                    End If
                End For
                PA = min(PartitionIndex, PA)
        End For
        Return PA
End Function
```